\documentclass{article}
\usepackage{spconf,amsmath,graphicx}
\usepackage{booktabs}

\usepackage{hyperref}
\title{National-scale 1-m resolution land-cover mapping for the entire China based on a low-cost solution and open-access data}
\name{Zhuohong Li, Wei He, Hongyan Zhang \thanks{Cossponding author: zhanghongyan@whu.edu.cn}}
\address{\small State Key Laboratory of Information Engineering in Surveying, Mapping and Remote Sensing, Wuhan University, Wuhan 430079, China}

\usepackage{enumitem}
\usepackage{subfigure}
\begin{document}
%
\maketitle
\begin{abstract}
Nowadays, many large-scale land-cover (LC) products have been released, however, current LC products for China either lack a fine resolution or nationwide coverage. With the rapid urbanization of China, there is an urgent need for creating a very-high-resolution (VHR) national-scale LC map for China. In this study, a novel 1-m resolution LC map of China covering $9,600,000 km^2$, called SinoLC-1, was produced by using a deep learning framework and multi-source open-access data. 
To efficiently generate the VHR national-scale LC map,
firstly, the reliable LC labels were collected from three 10-m LC products and Open Street Map data. Secondly, the collected 10-m labels and 1-m Google Earth imagery were utilized in the proposed low-to-high (L2H) framework for training. With weak and self-supervised strategies, the L2H framework resolves the label noise brought by the mismatched resolution between training pairs and produces VHR results. Lastly, we compare the SinoLC-1 with five widely used products and validate it with a sample set including 10,6852 points and a statistical report collected from the government. The results show the SinoLC-1 achieved an OA of 74\% and a Kappa of 0.65. Moreover, as the first 1-m national-scale LC map for China, the SinoLC-1 shows overall acceptable results with the finest landscape details.
\end{abstract}
\section{Introduction}
Land-cover (LC) mapping, as a basic earth observation application, provides an avenue for comprehending the land surface.
Over the past decades, China has undergone a tremendous wave of urbanization. Producing a VHR national-scale LC map of China can help us to investigate the environment, development, and future trend of the country in detail. With the updating of remote sensing platforms, the available LC products have been through the trends of coarse to fine \cite{wang2021loveda}. Nevertheless, due to the low orbit and small visual field of the VHR image captured platforms, the corresponding LC products generally have a smaller coverage when they have a higher spatial resolution. The published LC products which fully or partially covered China can be summarized in the following three general types:

\textbf{1)  Global-scale low-resolution products:}

From the 1980s to the 2000s, global-scale imagery with low resolution can be captured by SPOT 4, MODIS, and ENVISAT missions. Subsequently, many global land-cover (GLC) products, e.g., 300-m GlobCover, have emerged \cite{defourny2006globcover}.

\textbf{2) Global-scale moderate/high-resolution products:}
\begin{figure*}[t]
\centering
\includegraphics[width=1\linewidth]{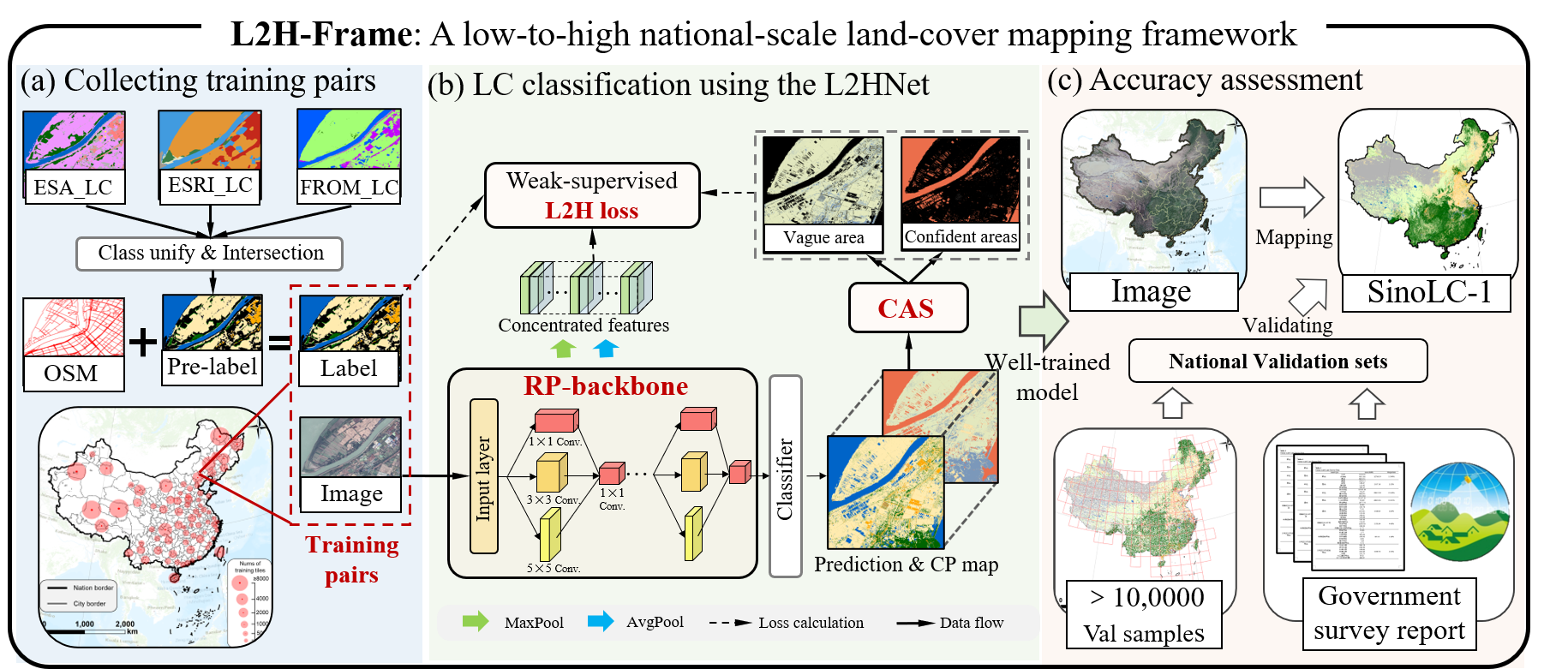}
\caption{ \footnotesize\rmfamily Overall workflow of the L2H-Frame. The framework includes three main parts: (a) Collecting training pairs; (b) Land-cover classification using the L2HNet; (c) Accuracy assessment.} 
\label{framework}
\end{figure*}

From the 2010s to 2020s, owing to the available Landsat and Sentinel imagery with moderate ($\sim$30m) and high ($\sim$10m) resolution, the related research has blossomed. E.g., GlobeLand30 \cite{chen2015global}, FROM\_GLC10 \cite{chen2019stable}, ESA\_WorldCover \cite{van2021esa}.

\textbf{3) Region-scale very-high-resolution datasets:}

In the 2020s, creating VHR datasets for deep learning research has become a hotspot, and current VHR LC datasets
(e.g., 2.1-m Hi-ULCM and 2.4-m PKU-USED) for China are
generally regional-scale (typically covering a few cities) \cite{huang2020high}.

Moreover, with the blossoming of deep learning techniques, deep learning has gradually become the predominant strategy conducted in VHR applications \cite{2021Outcome}. However, most of the approaches require a laborious annotation to produce accurate labels for training deep learning models, which inevitably limits them to small-scale applications.

In this study, we present a deep learning framework to create the 1-m LC map for China, called SinoLC-1, by using free access 1-m Google Earth imagery, 10-m GLC products, and Open Street Map (OSM) as input data. In detail, multi-source 10-m GLC products and OSM data are integrated to generate the coarse labels. Then, about 30\% areas of China are selected to train the low-to-high network (L2HNet), which was proposed in our previous work \cite{LI2022244}. By combining a multi-scale backbone, a weakly supervised module, and a self-supervised loss function in the L2HNet, the labeled noise caused by the mismatched resolution between VHR images and coarse labels is resolved during the training. Finally, the mapping results are predicted by the well-trained L2HNet using the nationwide VHR image batches and then merged into the seamless product. Moreover, the SinoLC-1 is produced without any commercial data and any manual annotations, which means it maintains both low capital expenditure and labor cost. To the best of our knowledge, the SinoLC-1 is the first 1-m resolution, which is also currently the highest resolution, LC product that covers the entire China.
\begin{figure*}[t]
\centering
\includegraphics[width=0.9\linewidth]{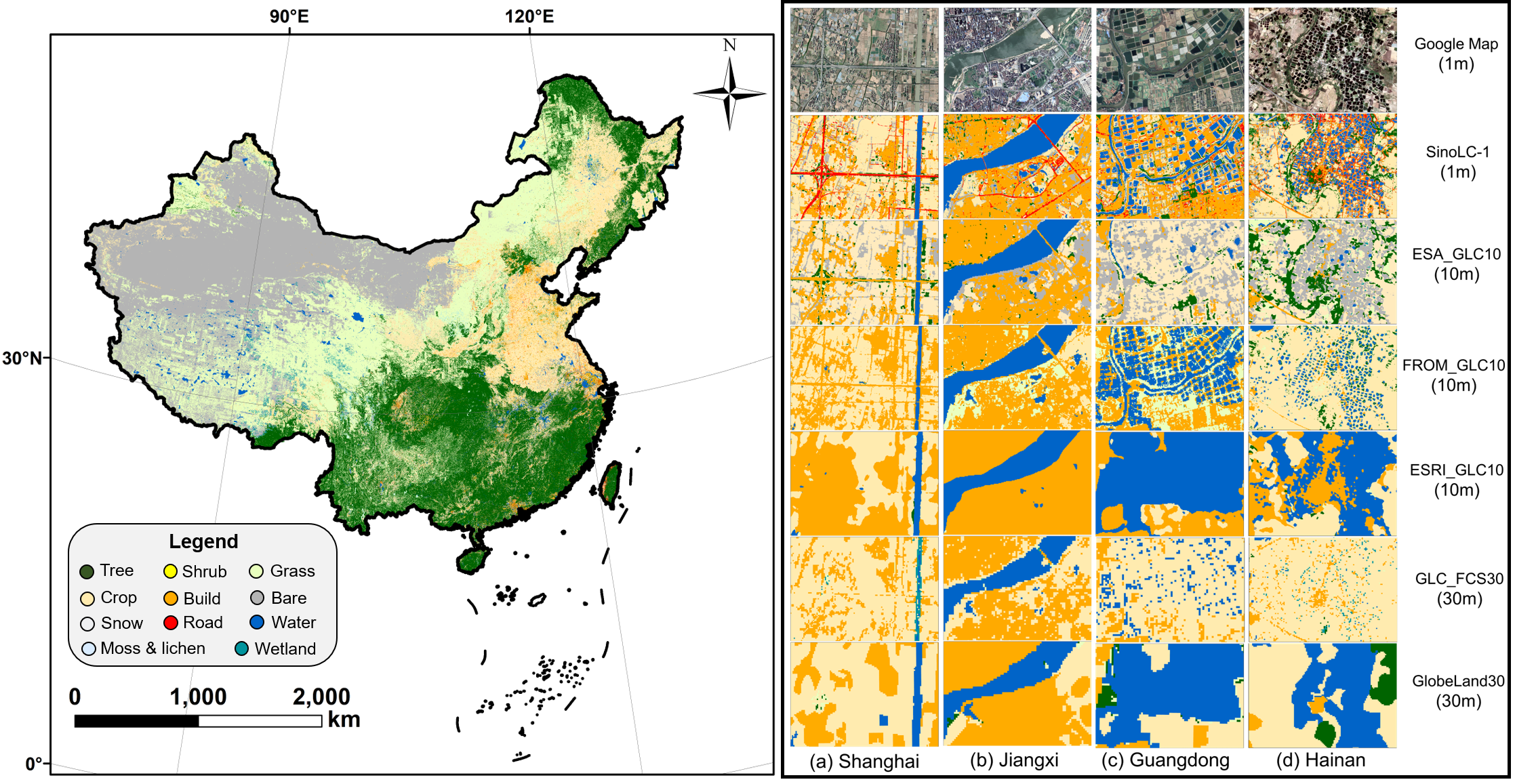}
\caption{ \footnotesize\rmfamily Demonstration of the SinoLC-1 and comparsion with other LC products.} 
\label{results}
\end{figure*}
\section{methodology}

\subsection{Collecting nationwide training pairs}

To collect reliable training data, as shown in Fig.\ref{framework}(a), 30\% of the areas were selected from the entire 34 provinces in China. Then, to obtain the stable LC information and generate the training labels, three 10-m GLC products (abbreviated as ESA\_GLC10 \cite{van2021esa}, ESRI\_GLC10 \cite{karra2021global}, and FROM\_GLC10 \cite{chen2019stable}) were unified according to their LC classes, and then the unified results were intersected to generate the pre-labels. In the pre-labels, the pixels, where their LC types were the same in the three GLC products, would be preserved as the stable labeled areas, otherwise, the pixels would be set as unlabeled type. Moreover, to generate stable labels for the traffic route, the vector road pattern information collected from the OSM was transferred to rasters with the same resolution as the pre-labels and then overlayed to the pre-labels to generate the final training labels. Finally, the training labels and the 1-m images form the training pairs and are sent to the network for training.

\subsection{LC mapping with the low-to-high network}

To process the training pairs, a low-to-high network (L2HNet) was applied. As shown in Fig.\ref{framework}(b), the L2HNet was designed by combining a resolution-preserving (RP) backbone, a weakly supervised-based confident areas selection (CAS) module, and a self-supervised-based low-to-high (L2H) loss. 

To robustly extract VHR features, the RP backbone consists of five blocks where each block contains multi-scale (1×1, 3×3, and 5×5) convolution layers with the channel setting of (64:32:16). Unlike the general encoder-decoder structures that deeply down-sample the features, the RP backbone scans the feature maps with properly small receptive fields and highly preserves their resolution. Lastly, the extracted features were classified into the prediction results with confidence probability (CP) maps.

To obtain reasonable supervision sources from the coarse label, the L2H loss was designed as a two-part composition. Firstly, considering the noisy samples of coarse labels, the CAS module selected the trustworthy parts from the labels according to the prediction CP map. Then, the confident area set, represented as \textbf{CA}, was selected to calculate the cross entropy (CE) loss with the coarse labels, and the vague area set with low confidence probability, represented as \textbf{VA}, was ignored during the CE calculation. Formally, for a training patch with the size of $W \times H$, we used ${\bf{Y'}}$,${\bf{\hat Y}}$, and $\bf{\hat G}$ to represent the labels, prediction results, and selected binary mask generated by the CAS module, respectively. The modified CE loss can be written as:
\begin{equation} 
\small
{\mathcal{L}_{CE}}({\bf{Y'}},{\bf{\hat Y}},{\bf{\hat G}}) = \frac{{ - \sum\limits_{i=0}^W {\sum\limits_{j=0}^H {\left[ {{{\hat g}_{ij}}\sum\limits_{l = 1}^L {{{y'}_{ij}}^{(l)}\log ({{\hat y}_{ij}}^{(l)})} } \right]} } }}{{card({\bf{CA}})}},
\label{CAS_LOSS}
\end{equation}
In Eq. \ref{CAS_LOSS}, ${y'_{ij}}^{(l)}$ and ${\hat y_{ij}}^{(l)}$ denote class $l$ of the 10-m label ${\mathbf{Y'}}$ and the prediction ${\mathbf{\hat Y}}$ in coordinates $(i,j)$, respectively.

Secondly, considering the feature similarity of the same LC classes, the dynamic vague area (DVA) loss was designed to constrain the within-class variance between the well-predicted \textbf{CA} and \textbf{VA} in the feature space. Formally, we used the 2-norm of the inter-area mean difference, represented as: $\sigma _{l,b}^2$, to describe the LC class $l \in [1,L]$ variance in the $b \in [1,B]$ feature layer of RP backbone. Moreover, the DVA loss is the accumulation of $\sigma _{l,b}^2$ in every LC class and feature layer, whose specific form is:
\begin{equation} \small
{\mathcal{L}_{DVA}} = \gamma \sum_{b=1}^{B} \sum_{l=1}^{L} {\sigma _{l,b}^2},
\label{DVA_LOSS}
\end{equation}
where $\gamma=0.05$ is a scale factor. By combining the Eqs. \ref{CAS_LOSS} and \ref{DVA_LOSS}, the L2H loss can be described as:
\begin{equation} \small
\mathcal{L}_{L2H} = {\mathcal{L}_{CE}}({\bf{Y'}},{\bf{\hat Y}},{\bf{\hat G}}) + \mathcal{L}_{DVA}.
\label{Total_LOSS}
\end{equation}
\begin{table*}[]
\caption{ \footnotesize\rmfamily {The confusion matrix for the SinoLC-1 land-cover product according to the national validation sample sets.}}
\centering
\footnotesize
\begin{tabular}{ccccclcccccccccc}
\hline
\multicolumn{2}{c}{LC type}          & TR          & TC           & \multicolumn{2}{c}{SL}                  & GL           & CL           & BD           & BL\&SV        & S\&I        & WT          & WL          & M\&L        & Total        & P.A. (\%)       \\ \hline
\multicolumn{2}{c}{TR}               & 447         & 173          & \multicolumn{2}{c}{5}                   & 209          & 184          & 228          & 240           & 0           & 28          & 0           & 0           & 1514         & 29.52           \\
\multicolumn{2}{c}{TC}               & 37          & 20708        & \multicolumn{2}{c}{14}                  & 2713         & 1899         & 124          & 134           & 0           & 352         & 5           & 52          & 26038        & 79.53           \\
\multicolumn{2}{c}{SL}               & 0           & 25           & \multicolumn{2}{c}{270}                 & 74           & 27           & 2            & 102           & 0           & 1           & 0           & 0           & 501          & 53.89           \\
\multicolumn{2}{c}{GL}               & 9           & 1332         & \multicolumn{2}{c}{35}                  & 17256        & 1837         & 119          & 2848          & 0           & 75          & 11          & 401         & 23923        & 72.13           \\
\multicolumn{2}{c}{CL}               & 53          & 1310         & \multicolumn{2}{c}{45}                  & 1976         & 11424        & 275          & 857           & 0           & 119         & 16          & 0           & 16075        & 71.07           \\
\multicolumn{2}{c}{BD}               & 57          & 83           & \multicolumn{2}{c}{3}                   & 72           & 274          & 1128         & 122           & 0           & 8           & 0           & 0           & 1747         & 64.57           \\
\multicolumn{2}{c}{BL\&SV}           & 50          & 209          & \multicolumn{2}{c}{23}                  & 5643         & 1031         & 418          & 24546         & 3           & 93          & 1           & 194         & 32211        & 76.20           \\
\multicolumn{2}{c}{S\&I}             & 0           & 2            & \multicolumn{2}{c}{0}                   & 94           & 7            & 0            & 51            & 135         & 2           & 0           & 92          & 383          & 35.25           \\
\multicolumn{2}{c}{WT}               & 2           & 21           & \multicolumn{2}{c}{0}                   & 39           & 105          & 12           & 59            & 0           & 1493        & 1           & 2           & 1734         & 86.10           \\
\multicolumn{2}{c}{WL}               & 0           & 37           & \multicolumn{2}{c}{11}                  & 46           & 28           & 3            & 7             & 0           & 14          & 135         & 0           & 281          & 48.04           \\
\multicolumn{2}{c}{M\&L}             & 0           & 22           & \multicolumn{2}{c}{2}                   & 698          & 18           & 2            & 455           & 2           & 5           & 0           & 733         & 1937         & 37.84           \\ \hline
\multicolumn{2}{c}{Total}            & 655         & 23922        & \multicolumn{2}{c}{408}                 & 28820        & 16834        & 2311         & 29421         & 140         & 2190        & 169         & 1474        & 106344       &                 \\
\multicolumn{2}{c}{U.A. (\%)}        & 6824        & 86.56        & \multicolumn{2}{c}{66.00}               & 59.88        & 67.86        & 48.81        & 83.43         & 96.43       & 68.17       & 79.88       & 49.73       &              &                 \\
\multicolumn{2}{c}{O.A. (\%)}        & \multicolumn{14}{c}{73.61}                                                                                                                                                                                                 \\
\multicolumn{2}{c}{Kappa}            & \multicolumn{14}{c}{0.6595}                                                                                                                                                                                                \\ \hline
Note:            & \multicolumn{15}{c}{\begin{tabular}[c]{@{}c@{}}TR=Traffic route; TC=Tree cover; SL=Shrubland; GL+BL\&SV=Grassland and Barren \&sparse vegetation; \\ CL=Cropland; BD=Building; S\&I=Snow \& ice;  WT=Water; M\&L=Moss \& lichen.\end{tabular}} \\ \hline
\end{tabular}
\label{cm}
\end{table*}

\section{Experiments}
\subsection{Validation set for accuracy assessment}
To comprehensively validate the quality of SinoLC-1, and to analyze the omission/commission error, as shown in Fig. \ref{framework}(c), we built a nationwide sample set by randomly sampling and visually interpreting 10,6852 points across China. Moreover, we derived a statistical validation set for every provincial administrative region by collecting the official third national land resource survey ($3^{\text{rd}}$ NLRS) data from the natural resources and planning bureau of the Chinese government.

\subsection{Qualitative comparison with other LC products}

Intuitively, based on the established knowledge, the overall result of SinoLC-1 shown in Fig. \ref{results} accurately reflects the geospatial distribution of multiple LC types and highly conforms to the actual LC pattern of China.  
In detail, we illustrated four typical regions as demonstration areas to compare the SinoLC-1 with five widely used large-scale LC products, which were mentioned in Section 1. By comparing the urban areas shown in Fig. \ref{results}(a)(b), the SinoLC-1 provided more accurate information on slender roads and performed better in indicating the finer urban pattern. By comparing the agricultural areas shown in Fig. \ref{results}(c)(d), SinoLC-1 indicates the most accurate LC situations, where all single ponds are predicted.

\subsection{Quantitative analysis and accuracy assessment}
\begin{figure}[]
{
    \begin{minipage}[b]{\hsize}
     \centering
    \includegraphics[width=0.9\linewidth]{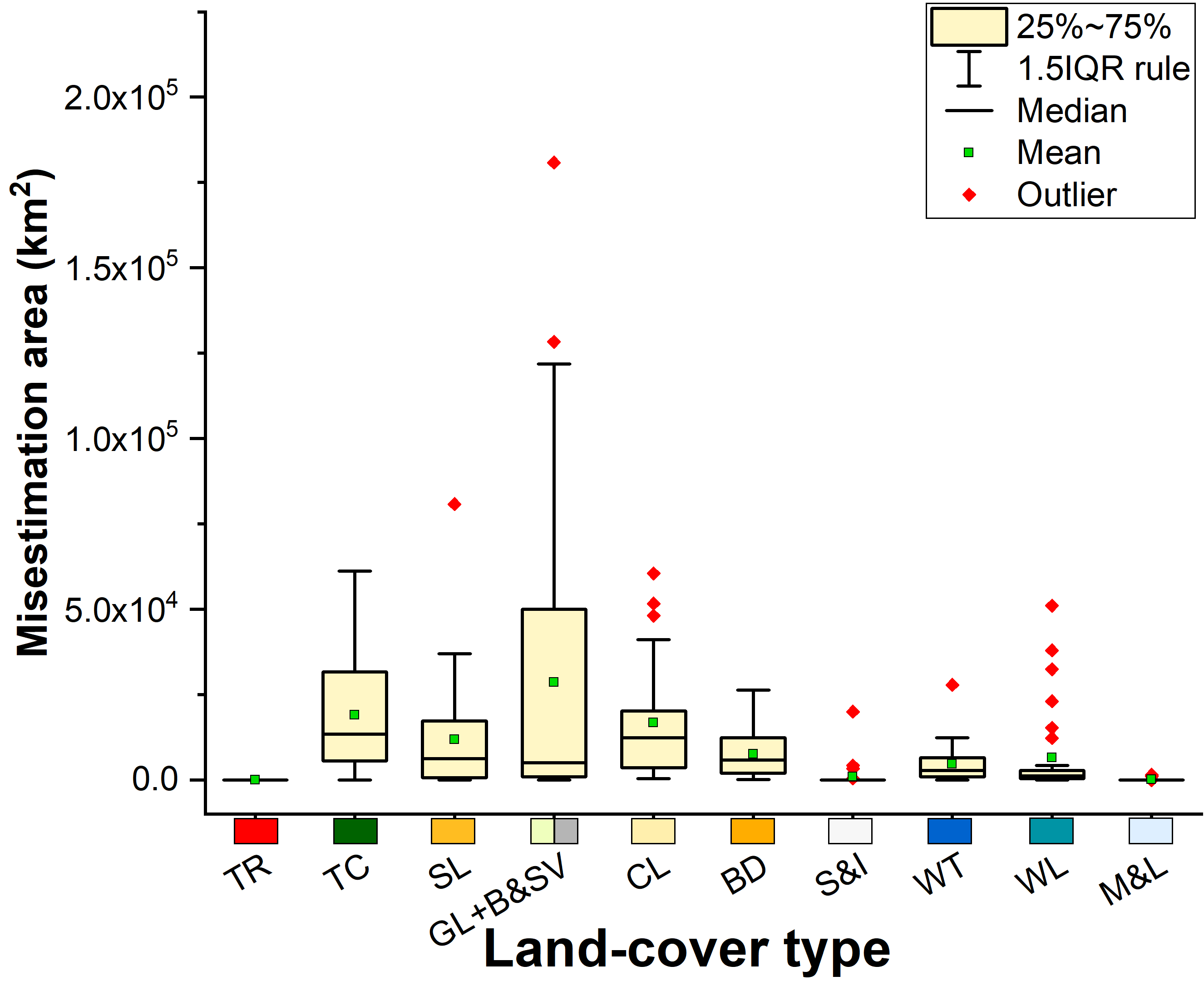}
    \end{minipage}
    } 
    \vspace{-0.2cm}
\caption{ \footnotesize\footnotesize \rmfamily Misestimation in every land-cover type through 31 provinces in China (Macao, Hong kong, and Taiwan are not available in $3^{\text{rd}}$ NLRS data).}
\label{sta}
\end{figure}

From the confusion matrix shown in Table \ref{cm}, the SinoLC-1  achieved an OA of 73.61\% and a kappa coefficient of 0.6595. The results show that the basic LC types with large coverage (i.e., tree, cropland, water, etc) had higher accuracies since they generally have distinguishable features and occupied large proportions. 
To evaluate the overall misestimation distributions of SinoLC-1, based on the government reports collected from $3^{\text{rd}}$ NLRS, we described the error distribution of every LC type in 31 provincial administrative regions. From Fig. \ref{sta}, although there are some outliers in “grassland” and “barren and sparse vegetation” types, which occupied a very large proportion in northwest China, the low misestimation level of overall results indicates that the SinoLC-1 is a statistically acceptable LC product across China. 

\section{Conclusion}

In this study, a 1-m national-scale LC product for China was produced based on a low-cost L2H-Frame and free access data. 
The qualitative comparison with five GLC products shows the SinoLC-1 has the finest LC details and better portrays the dense urban pattern. The quantitative validation indicates the SinoLC-1 achieved an OA of 73.61\% and statistically conforms to the official survey reports in most of the LC types and regions across China. 
Overall, it concludes the SinoLC-1 is a promising and accurate product to provide clear information and vital support for China. To the best of our knowledge, the produced SinoLC-1 is the first 1-m LC product that can cover the entire China. In the future, we will continually optimize and update the SinoLC-1 for providing more accurate and sustained LC information.

\section{Acknowledges}

We acknowledge the free access LC products provided by ESA, ESRI, Inc., and Tsinghua University, the traffic route information provided by the OSM, and the VHR images provided by Google Inc.

\small
\bibliographystyle{IEEEbib}
\bibliography{strings,refs}

\begin{thebibliography}{1}

\bibitem{wang2021loveda}
Junjue Wang et~al.,
\newblock ``{LoveDA}: A remote sensing land-cover dataset for domain adaptive
  semantic segmentation,''
\newblock in {\em Thirty-fifth Conference on Neural Information Processing
  Systems Datasets and Benchmarks Track (Round 2)}, 2021.

\bibitem{defourny2006globcover}
Pierre Defourny et~al.,
\newblock ``{GLOBCOVER}: a 300 m global land cover product for 2005 using
  envisat meris time series,''
\newblock in {\em Proceedings of ISPRS Commission VII Mid-Term Symposium}.
  Citeseer, 2006, pp. 8--11.

\bibitem{chen2015global}
Jun Chen et~al.,
\newblock ``Global land cover mapping at 30 m resolution: A pok-based
  operational approach,''
\newblock {\em ISPRS Journal of Photogrammetry and Remote Sensing}, vol. 103,
  pp. 7--27, 2015.

\bibitem{chen2019stable}
Bin Chen et~al.,
\newblock ``Stable classification with limited sample: Transferring a 30-m
  resolution sample set collected in 2015 to mapping 10-m resolution global
  land cover in 2017,''
\newblock {\em Sci. Bull}, vol. 64, pp. 370--373, 2019.

\bibitem{van2021esa}
Ruben Van De~Kerchove et~al.,
\newblock ``{ESA WorldCover:} global land cover mapping at 10 m resolution for
  2020 based on sentinel-1 and 2 data.,''
\newblock in {\em AGU Fall Meeting Abstracts}, 2021, vol. 2021, pp.
  GC45I--0915.

\bibitem{huang2020high}
Xin Huang et~al.,
\newblock ``High-resolution urban land-cover mapping and landscape analysis of
  the 42 major cities in china using zy-3 satellite images,''
\newblock {\em Science Bulletin}, vol. 65, no. 12, pp. 1039--1048, 2020.

\bibitem{2021Outcome}
Zhuohong Li et~al.,
\newblock ``The outcome of the 2021 {IEEE GRSS Data Fusion Contest—Track
  MSD:} multitemporal semantic change detection,''
\newblock {\em IEEE Journal of Selected Topics in Applied Earth Observations
  and Remote Sensing}, vol. 15, pp. 1643--1655, 2022.

\bibitem{LI2022244}
Zhuohong Li et~al.,
\newblock ``Breaking the resolution barrier: A low-to-high network for
  large-scale high-resolution land-cover mapping using low-resolution labels,''
\newblock {\em ISPRS Journal of Photogrammetry and Remote Sensing}, vol. 192,
  pp. 244--267, 2022.

\bibitem{karra2021global}
Krishna Karra et~al.,
\newblock ``Global land use/land cover with sentinel 2 and deep learning,''
\newblock in {\em 2021 IEEE international geoscience and remote sensing
  symposium IGARSS}. IEEE, 2021, pp. 4704--4707.

\end{thebibliography}

\end{document}